\documentclass[letterpaper, 10 pt, conference]{ieeeconf}  

\IEEEoverridecommandlockouts                              

\usepackage{graphics} 
\usepackage{epsfig} 
\usepackage{mathptmx} 
\usepackage{times} 
\usepackage{amsmath} 
\usepackage{amssymb}  
\usepackage{algorithm}
\usepackage{siunitx}
\usepackage[dvipsnames]{xcolor}
\usepackage{algpseudocode}
\usepackage{array}
\usepackage{cite}
\usepackage{subcaption}

\newcommand{\compressParag}{\looseness=-1}
\title{\LARGE \bf
A Nonlinear Model Predictive Control for Automated Drifting with a Standard Passenger Vehicle*}

\author{Stan Meijer$^{1}$, Alberto Bertipaglia$^{2}$ and Barys Shyrokau$^{2}$
\thanks{*The Dutch Science Foundation NWO-TTW supports the research within the EVOLVE project (nr. 18484).}
\thanks{$^{1}$Stan Meijer is with the Department of Driving Dynamics, BMW Group, 80788 Munich, Germany. 
        {Stan.Meijer@bmw.de}}%
\thanks{$^{2}$Alberto Bertipaglia and Barys Shyrokau are with the Department of Cognitive Robotics, Delft University of Technology, 2628 CD Delft, The Netherlands.
        {\{A.Bertipaglia, B.Shyrokau\}@tudelft.nl}}%
}

\begin{document}

\maketitle
\thispagestyle{empty}
\pagestyle{empty}

\begin{abstract}
This paper presents a novel approach to automated drifting with a standard passenger vehicle, which involves a Nonlinear Model Predictive Control to stabilise and maintain the vehicle at high sideslip angle conditions. The proposed controller architecture is split into three components. The first part consists of the offline computed equilibrium maps, which provide the equilibrium points for each vehicle state given the desired sideslip angle and radius of the path. The second is the predictive controller minimising the errors between the equilibrium and actual vehicle states. The third is a path-following controller, which reduces the path error, altering the equilibrium curvature path. In a high-fidelity simulation environment, we validate the controller architecture capacity to stabilise the vehicle in automated drifting along a desired path, with a maximal lateral path deviation of 1 m. In the experiments with a standard passenger vehicle, we demonstrate that the proposed approach is capable of bringing and maintaining the vehicle at the desired 30 deg sideslip angle in both high and low friction conditions. \compressParag
\end{abstract}

%
\section{Introduction}
\begin{figure}[!t] 
    \centering
    \includegraphics[width=\linewidth, keepaspectratio]{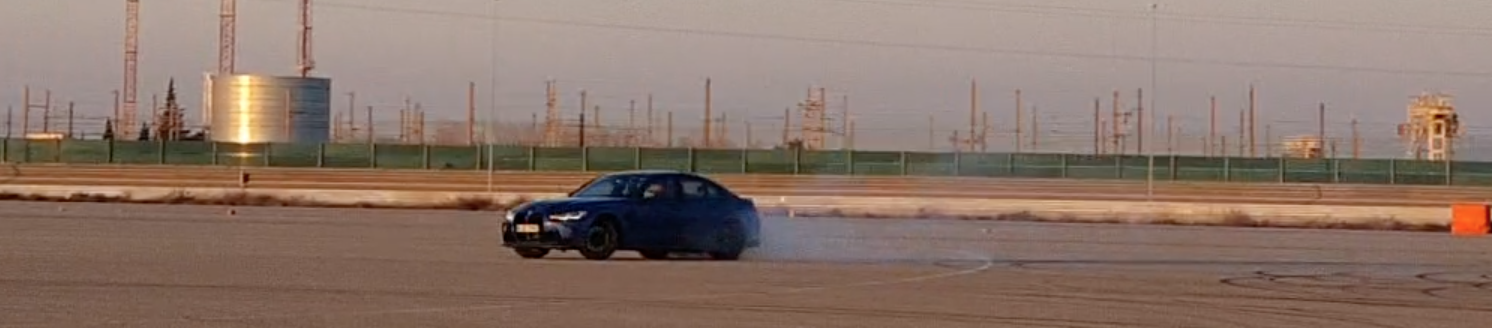}
    \caption{BMW M3 Competition in automated drifting.}
    \label{fig:Drifting}
\end{figure}

The ability to control the vehicle beyond the friction limit or at a high sideslip angle extends the number of controllable vehicle states and possible trajectories. Thus, it improves vehicle safety for the collision avoidance manoeuvres in which a conventionally driven vehicle would have reached its handling limit \cite{velenis2011steady, bertipaglia2023model, lenssen2023combined}. However, driving along a desired path while sustaining a large sideslip angle is particularly challenging because it requires exploiting the coupled nonlinearities in the tyre force response \cite{abdulrahim2006dynamics}. Furthermore, it is necessary to keep the rear tyres saturated \cite{hindiyeh2014controller}. Thus, in this paper, we focus on designing a nonlinear model predictive control (NMPC), which can stabilise a vehicle in a drifting motion while remaining on a desired path. Furthermore, we aim to implement and evaluate the proposed controller in a real-world experiment on a standard passenger vehicle without hardware modification (Fig. \ref{fig:Drifting}). \compressParag

Different control techniques for automated drifting have recently been proposed in the literature. A possible solution is applying a linear quadratic regulator (LQR) controller based on the single-track model combined with a Fiala tyre model \cite{park2021experimental}. The control inputs are the steering angle and rear wheel speed, computing from the required rear longitudinal force component. The internal model has also been upgraded with a double-track vehicle model based on a Pacejka tyre model, to reduce the model mismatches \cite{velenis2011steady}. However, all the proposed LQR controllers only stabilise the vehicle in the drifting equilibrium while an external path-tracking controller executes the path following control. \compressParag

Thus, other approaches are proposed that simultaneously ensure the path-following properties and control the vehicle into a desired equilibrium drifting state  \cite{goh2016simultaneous, goh2020toward}. These controllers use the lateral error with respect to a reference path to control the vehicle along the desired trajectory. At the same time, the sideslip error with respect to a reference sideslip angle brings the vehicle into a state of drifting. The steering angle and rear drive torque are treated as control inputs, and they are determined through a function of imposed error dynamics, a nonlinear model inversion followed by a wheel speed control. The imposed error dynamics convert the lateral and sideslip errors into a desired yaw rate and yaw acceleration. A numerical approach of nonlinear model inversion is applied to determine the desired steering and throttle angles. The steering angle is directly applied to the system, while the throttle angle is mapped to a desired wheel speed. Experimental verification of these controllers has shown that the approach can successfully track a reference path and sideslip with good performance. However, the vehicle powertrain layout includes independent electric engines simulating the behaviour of a fully locked differential, which is uncommon for standard passenger vehicles. \compressParag

Several solutions based on Model Predictive Control (MPC) are proposed in the literature \cite{Czibere2021ModelSurfaces, zhang2018tire, Beal2013ModelHandling}. For instance, a simplified linear vehicle can be implemented as an MPC prediction model  \cite{Beal2013ModelHandling}. The MPC modifies a human driver input such that a vehicle is stabilised on the handling limits. As modified linear vehicle and tyre models are used, the optimisation becomes convex, reducing the computational effort and increasing real-time feasibility. However, the controller is not designed to bring a vehicle into a steady-state drift fully. Thus, another solution is linearising the nonlinear single-track vehicle with a Fiala tyre model around defined state and input variables  \cite{Czibere2021ModelSurfaces}. As a steady-state drift requires a reference equilibrium state, the system will change according to the values of equilibria through equilibrium-varying simulation. The approach shows that the prediction model states can be chosen arbitrarily, as the sideslip angle is not chosen as a state, but tracking the reference drifting state is still possible. However, the approach still suffers from a computational point of view, making the controller not real-time feasible. Thus, in this paper, we develop a real-time NMPC based on a nonlinear single-track vehicle with a simplified Pacejka tyre model, using the sideslip angle as a vehicle state. The proposed controller aims to bring the vehicle to a high sideslip angle and stabilise the vehicle in this unstable equilibrium, considering high and low friction conditions. \compressParag

The main contribution is the experimental validation of the proposed NMPC for automated drifting with a standard passenger vehicle, contrary to previous works in the literature that are limited to simulations \cite{acosta2018full, stanoEnhanced2023} or experimental demonstrators with heavy hardware modifications \cite{goh2016simultaneous, goh2020toward}. \compressParag
\section{Vehicle Model}
Subsection \ref{sub:SingleTrack} describes the nonlinear single-track model, and subsection \ref{sub:Tyre} explains the tyre model.

\subsection{Single-Track Vehicle Model}
\label{sub:SingleTrack}
The single-track vehicle model simplifies the dynamics of a four-wheel vehicle model by lumping the tyres on the left and right sides together in the centre axis of the vehicle and ignoring the lateral load transfer. The kinematics of the single-track vehicle model is described using the Frenet reference system as follows:
\begin{equation}
    \begin{cases}
        \dot{\psi} &= r-K \frac{V \cos(\psi+\beta)}{1-e_y K}\\
        \dot{e_y} &= V \sin(\psi+\beta)\\
        \dot{s_x} &= \frac{V \cos(\psi+\beta)}{1-e_y K}
    \end{cases}
\end{equation}
where $\psi$ is the heading angle, $s_x$ is the travelled distance with respect to the desired path, and $e_y$ is the normal distance from the desired path. The relation between the absolute vehicle velocity $V$, path curvature $K$ and yaw rate $r$ describes the desired trajectory.

The dynamics of the nonlinear single-track vehicle model are defined in Eq.\ref{eq:bicycle}, where the vehicle states are the sideslip angle $\beta$, the yaw rate, and the absolute vehicle velocity. The absolute velocity describes the actual velocity of the vehicle in space, combining both the lateral and longitudinal velocity components with respect to the vehicle frame. The sideslip angle is the angle between the vehicle's longitudinal axis and the absolute velocity, which is, therefore, the main indication for a vehicle being in a drift at large sideslip angles \cite{bertipaglia2022model, bertipaglia2023unscented, Bertipaglia2022Two}. Using the absolute velocity, a path following property is established, as path curvature can be expressed as a function of absolute velocity divided by the yaw rate. The suspension dynamics are not modelled to reduce the computational effort, so the longitudinal weight transfer is considered in a quasi-static way.\compressParag
\begin{equation}
    \begin{cases}
        \dot{V} = \dfrac{1}{m}\big(F_{xF}\cos(\delta-\beta)-F_{yF}\sin(\delta-\beta)+\\
        \indent +F_{xR}\cos{\beta}+F_{yR}\sin{\beta}\big)\\
        \dot{\beta} = \dfrac{1}{m V}\big(F_{yF}\cos(\delta-\beta)-F_{xF}\sin(\delta-\beta)+\\
        \indent -F_{xR}\sin{\beta}+F_{yR}\cos{\beta}\big)-r\\
        \dot{r} = \dfrac{a \big(F_{xF}\sin{\delta}+F_{yF}\cos{\delta}\big)-b F_{yR}}{I_z}\\
    \end{cases}
    \label{eq:bicycle}
\end{equation}

\subsection{Tyre Model}
\label{sub:Tyre}
The vehicle and road surface interaction is essential for the drifting motion. The transition between conventional driving and drifting requires a large rear lateral force to establish a state of drift. Similarly, a high optimal lateral force must be maintained to remain in a steady-state drift. This implies that the knowledge of the tyres should be as accurate as possible, capturing most characteristics with the lowest complexity. This work combines the standard Magic Formula (5.2) and its simplified version. Results on computational capacity and tracking performance show that the simplified Magic Formula can be applied at the front wheels.
In contrast, the standard formula is desired to better capture rear tyre characteristics. The simplified model can capture those characteristics as front wheels are considered within the tyre friction limits during a drift. Regarding rear tyres, they work beyond the friction limit, so tyre characteristics should be captured as well as possible.
%
\section{Equilibrium Analysis}
\label{Sec:Stability}
Subsection \ref{sub:equilibriapts} describes how the locations of the desired steady-state drift equilibria are computed, and subsection \ref{sub:verifequilibria} shows the experimental validation of the drifting equilibria.
\begin{figure*}[!t]
    \centering
    \subcaptionbox*{(a) Equilibria map $F_{yF}^{eq}$}[.33\linewidth]{%
    \includegraphics[width=\linewidth]{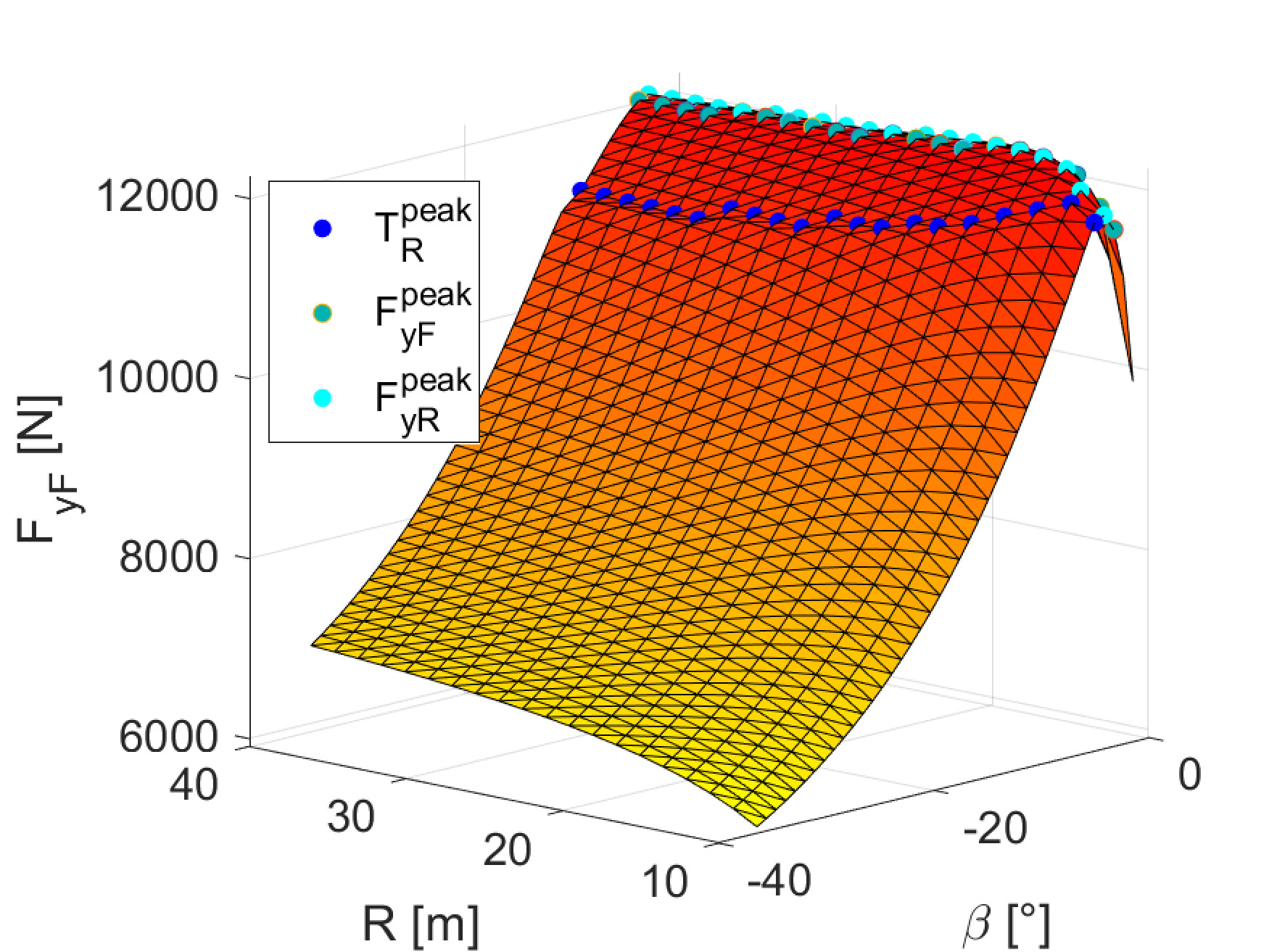}%
    }%
    \hfill
    \subcaptionbox*{(b) Equilibria map $F_{yR}^{eq}$}[.33\linewidth]{%
    \includegraphics[width=\linewidth]{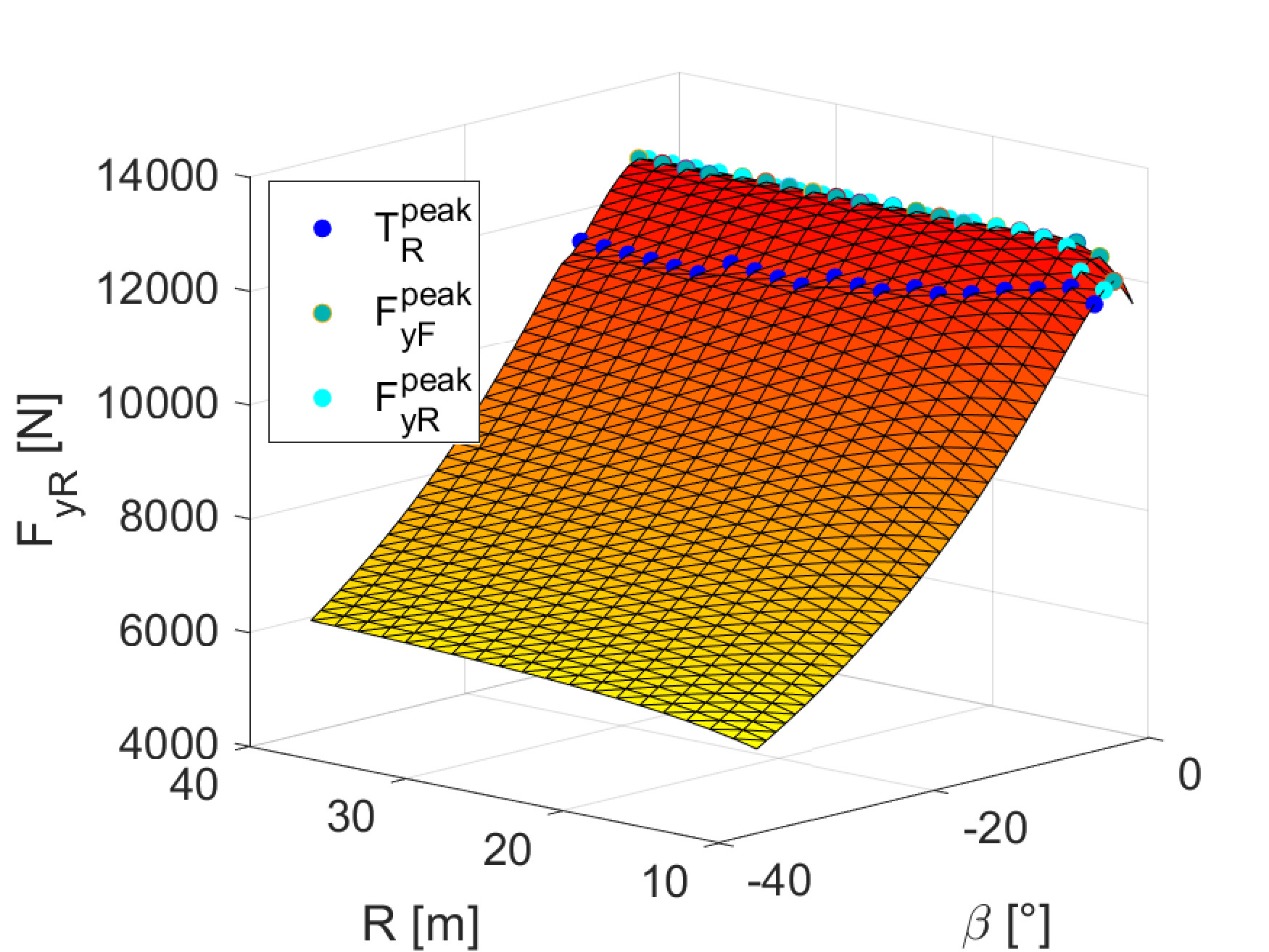}%
    }%
    \hfill
    \subcaptionbox*{(c) Equilibria map $\omega_{F}^{eq}$ and $\omega_{R}^{eq}$}[.33\linewidth]{%
    \includegraphics[width=\linewidth]{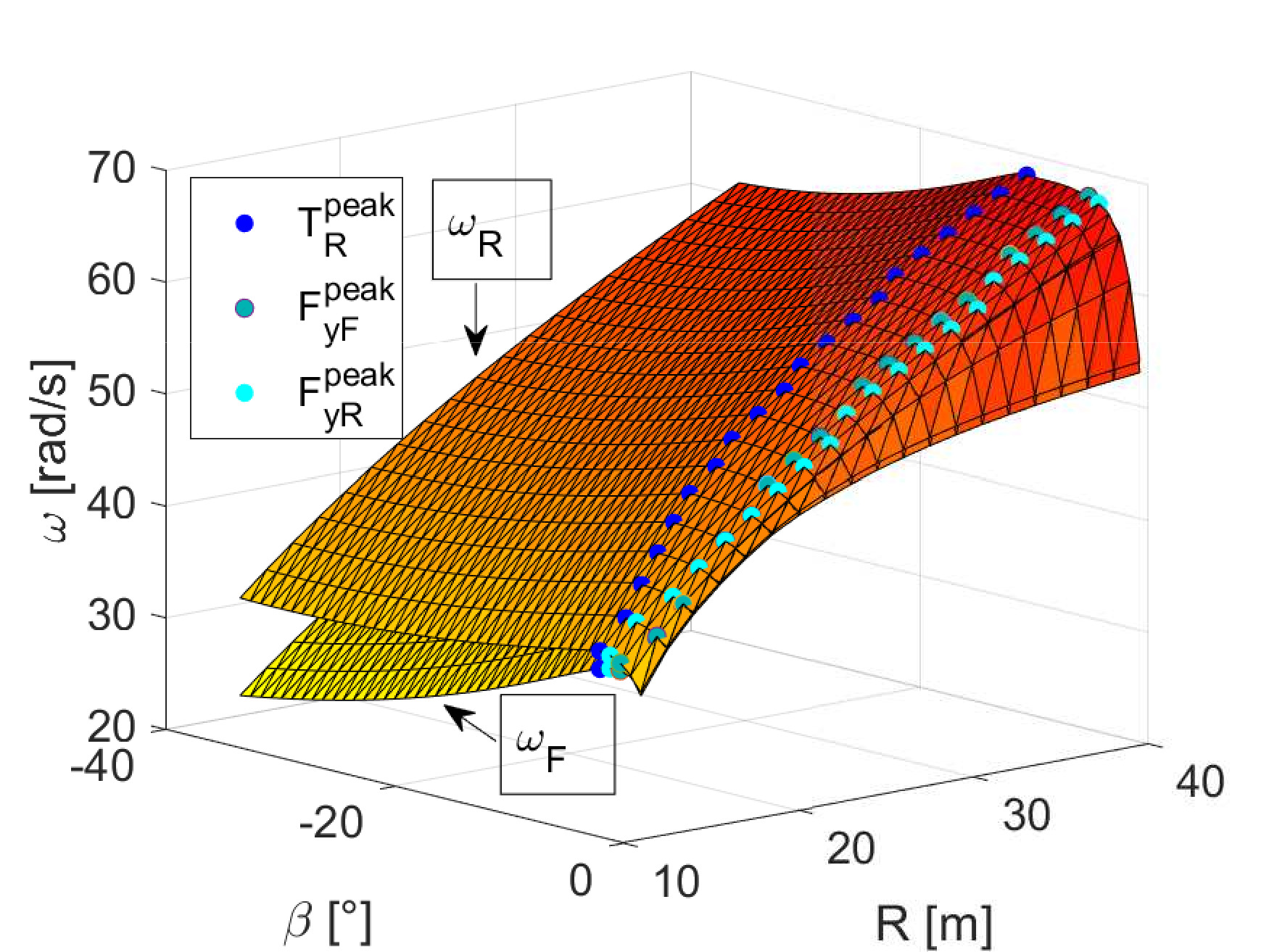}%
    }%
    \hfill\\
    \subcaptionbox*{(d) Equilibria map $T_{R}^{eq}$}[.33\linewidth]{%
    \includegraphics[width=\linewidth]{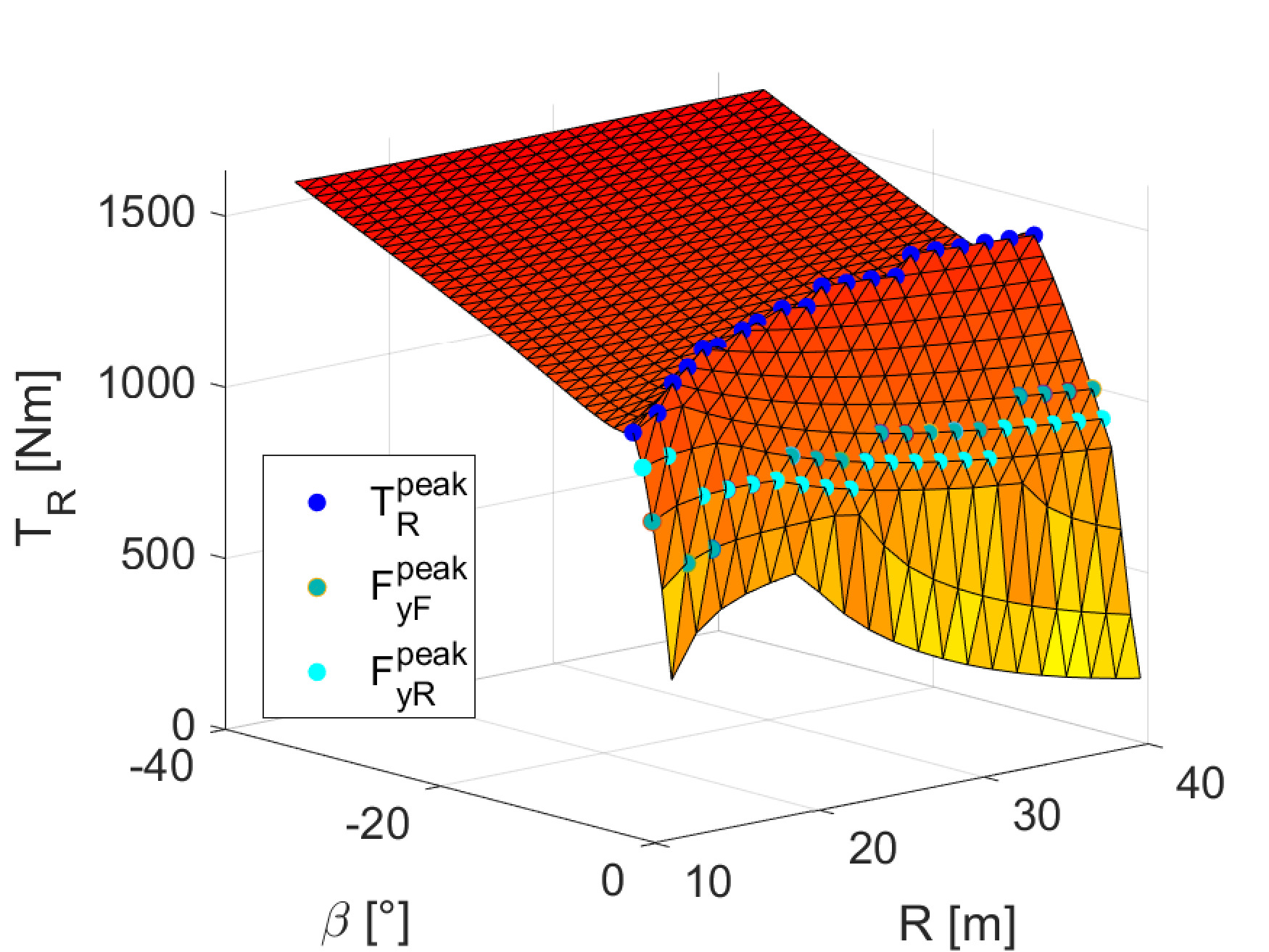}%
    }
    \hfill
    \subcaptionbox*{(e) Equilibria map $\delta^{eq}$}[.33\linewidth]{%
    \includegraphics[width=\linewidth]{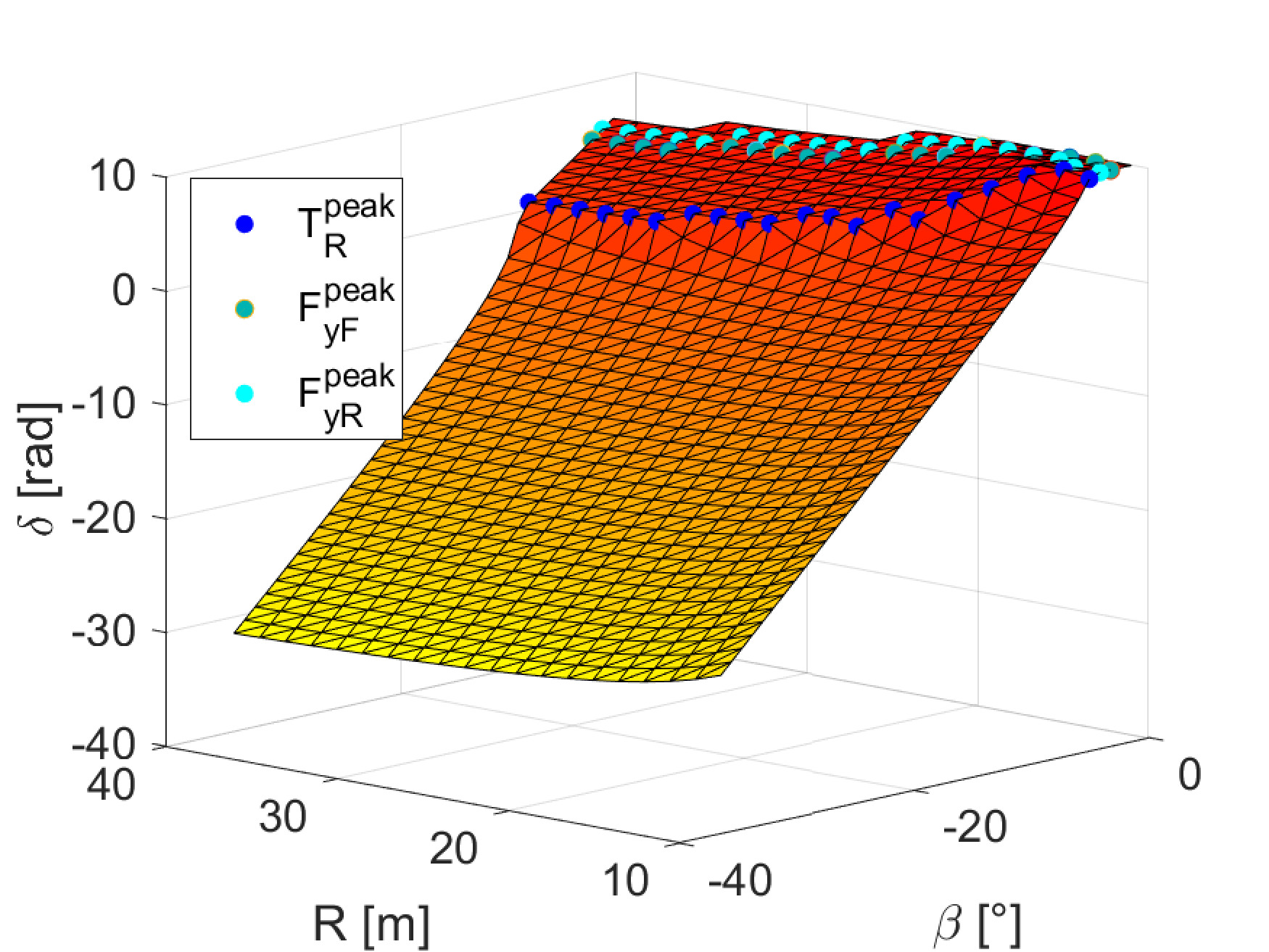}%
    }%
    \hfill
    \subcaptionbox*{(f) Equilibria map $V^{eq}$}[.33\linewidth]{%
    \includegraphics[width=\linewidth]{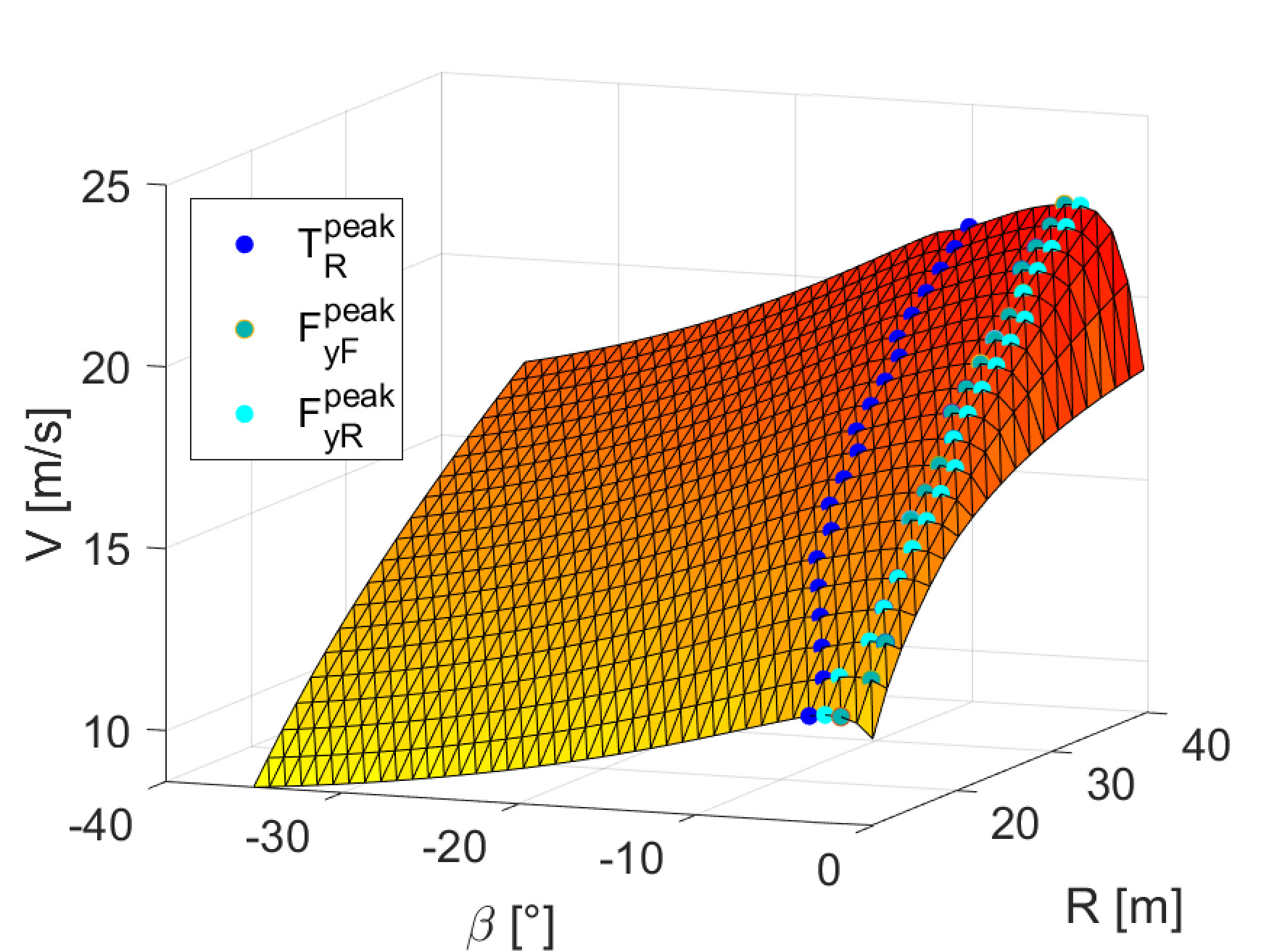}%
    }%
    \caption{Steady-state drifting equilibria maps.}
    \label{fig:Equilibrium}
\end{figure*} 

\subsection{Steady-State Equilibrium Locations}
\label{sub:equilibriapts}
The steady-state drifting equilibria are computed, imposing the derivative of the nonlinear single-track vehicle model states to zero. Based on this condition, the differential equations are solved using nonlinear least-squares, allowing the incorporation of constraints on respective states. An initial guess allows the solver to converge to a desired equilibrium, e.g. a high sideslip drifting equilibrium. Furthermore, constraints on desired velocities, wheel speeds, and wheel slip make finding equilibria significantly faster. Nevertheless, as can be mathematically derived, the proposed model contains more variables than equations, which means that specific equilibrium values must be user-defined, considering a value that describes a drifting motion. Thus, the system of equations is solved using a set of desired sideslip angles $\beta^{eq}$ with a set of experimentally confirmed feasible path radii $R^{eq}$ for drifting motion. The optimisation variables are described by $\textbf{x}_{eq}$ (Eq.\ref{eq:optimstates}), which allows the computation of all the respective tyre force components.\compressParag
\begin{equation}
    \begin{aligned}
        \textbf{x}^{eq} = [V^{eq} && \beta^{eq} && r^{eq} &&  \omega_F^{eq} && \omega_R^{eq} && \delta^{eq} && T_i^{eq}]^T
    \end{aligned}
    \label{eq:optimstates}
\end{equation}
The results of the equilibrium points are considered non-unique, as a drifting equilibrium is controlled through a balancing relation between the front and rear force magnitudes. Thus, the variations in rear-drive torques and their resultant slip coefficient may vary, as the front axle is subject to varying poses and forces. Fig. \ref{fig:Equilibrium} shows the equilibria maps, which have the locations of the peak drive torques in $\color{blue} \bullet$, rear lateral peak forces in $\color{cyan}\bullet$ and the front lateral peak forces in $\color{teal}\bullet$. 
For the velocity state equilibrium, see Fig. \ref{fig:Equilibrium}f, which increases with the desired path radius at a constant sideslip angle. In contrast, a constant radius causes a decrease in sideslip angle due to the force balance required between the front and rear axles. As a result, the vehicle's yaw rate increases with the path radius. Fig. \ref{fig:Equilibrium}a and \ref{fig:Equilibrium}b show that the lateral tyre forces are at their highest at relatively small sideslip angles. Thus, towards these peaks, the vehicle is considered in conventional driving with the tyre working in the linear range. This can also be deducted by the similar wheel speeds (Fig. \ref{fig:Equilibrium}c) and positive steering angles (Fig. \ref{fig:Equilibrium}e).
Fig. \ref{fig:Equilibrium}c shows that the front and rear wheel speeds start to deviate in magnitude at higher sideslip angles, and the steering angle approaches zero, indicating that the vehicle is entering a drifting mode. However, the rear longitudinal force has not reached its peak force (Fig. \ref{fig:Equilibrium}b). This means that the applied rear torque results in a reasonable amount of slip such that the rear tyres become saturated in the lateral direction with a reasonable sideslip angle. However, since the peak friction force in the longitudinal direction is not reached, we consider the state towards this point a transient equilibrium, i.e., a slight drift. Thus, tyres are not fully saturated, nor is countersteering taking place. 
Once the peak in the longitudinal slip is reached, the difference in wheel speeds becomes more significant than the rear axle longitudinal velocity. Thus, the tyres are saturated in both lateral and longitudinal directions, and a direct change in steering angle is required (Fig. \ref{fig:Equilibrium}e) to balance the front and rear forces. Thus, larger sideslip angles describe fully drifting equilibria where the proportional behaviour in the steering angle and the deviation in the wheel speeds (Fig. \ref{fig:Equilibrium}c) result from a more significant required rear slip and a decline in front wheel speed due to a lower yaw rate.

\subsection{Experimental Validation}
\label{sub:verifequilibria}
\begin{figure}[!t]
    \centering
    \includegraphics[width=1\linewidth, keepaspectratio]{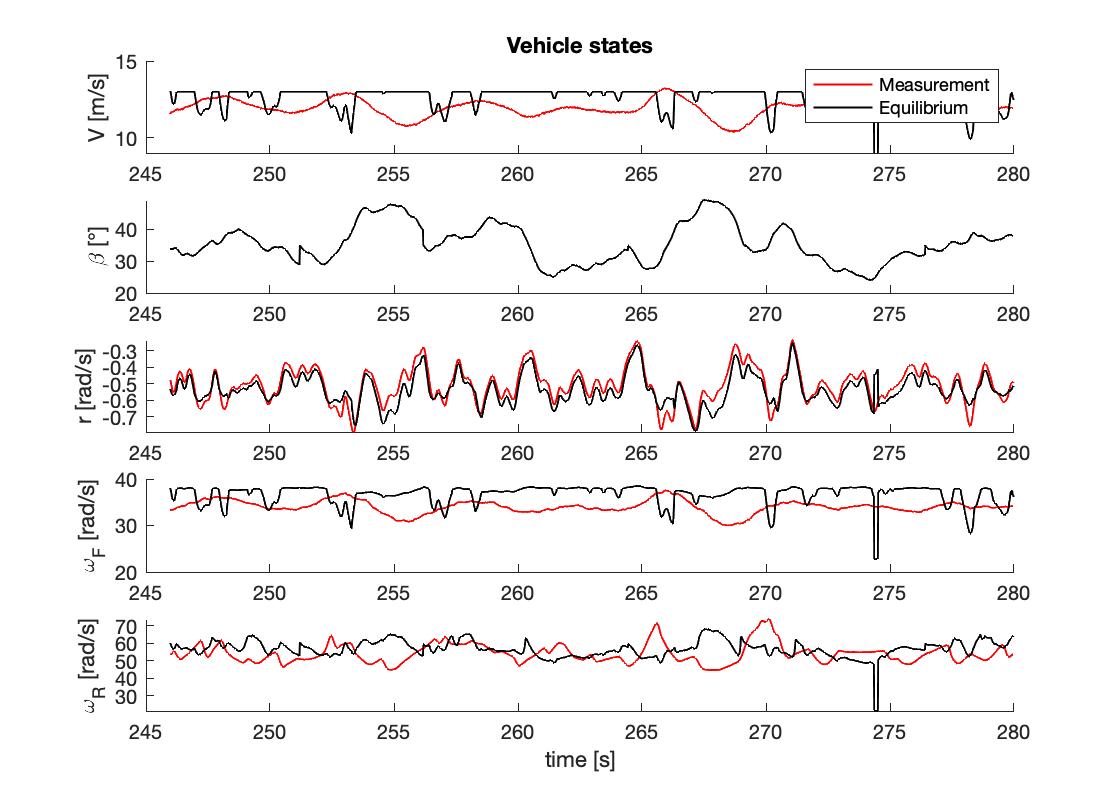}
    \caption{Comparison of the computed and measured vehicle states in equilibrium.}
    \label{fig:eqcompstate}
\end{figure}

The steady-state drifting equilibria are experimentally validated in a proving ground. Fig. \ref{fig:eqcompstate} shows a comparison of the computed and measured vehicle state equilibria. The results show the complexity of a human driver remaining at an actual equilibrium point due to many disturbances on the vehicle, e.g., road surface. Real road surfaces are considered non-homogeneous, where a \SI{\sim5}{\%} variation in the friction coefficient is measured for the respective proving ground, which a driver needs to compensate for by changing the steering angle or throttle commands. Thus, the measured steering angle and drive torque are never constant.
The vehicle's slightly tilted position due to the surface's conical shape is also unmodelled in the single-track vehicle, causing variations in the vertical axle loads. In addition, the driver has to stabilise the desired drifting state while stabilising on a desired path. Thus, the experimental data are not in equilibrium for all time instants. However, the comparison does show reasonable similarities in the order of magnitude. 
The comparison between computed and measured drifting states shows that many variations of drifting equilibria exist around an actual drift equilibria. 
\section{Proposed Controller Architecture}
\label{sec:Controller}
\begin{figure}[!t]
    \centering
    \includegraphics[width=1\linewidth, keepaspectratio]{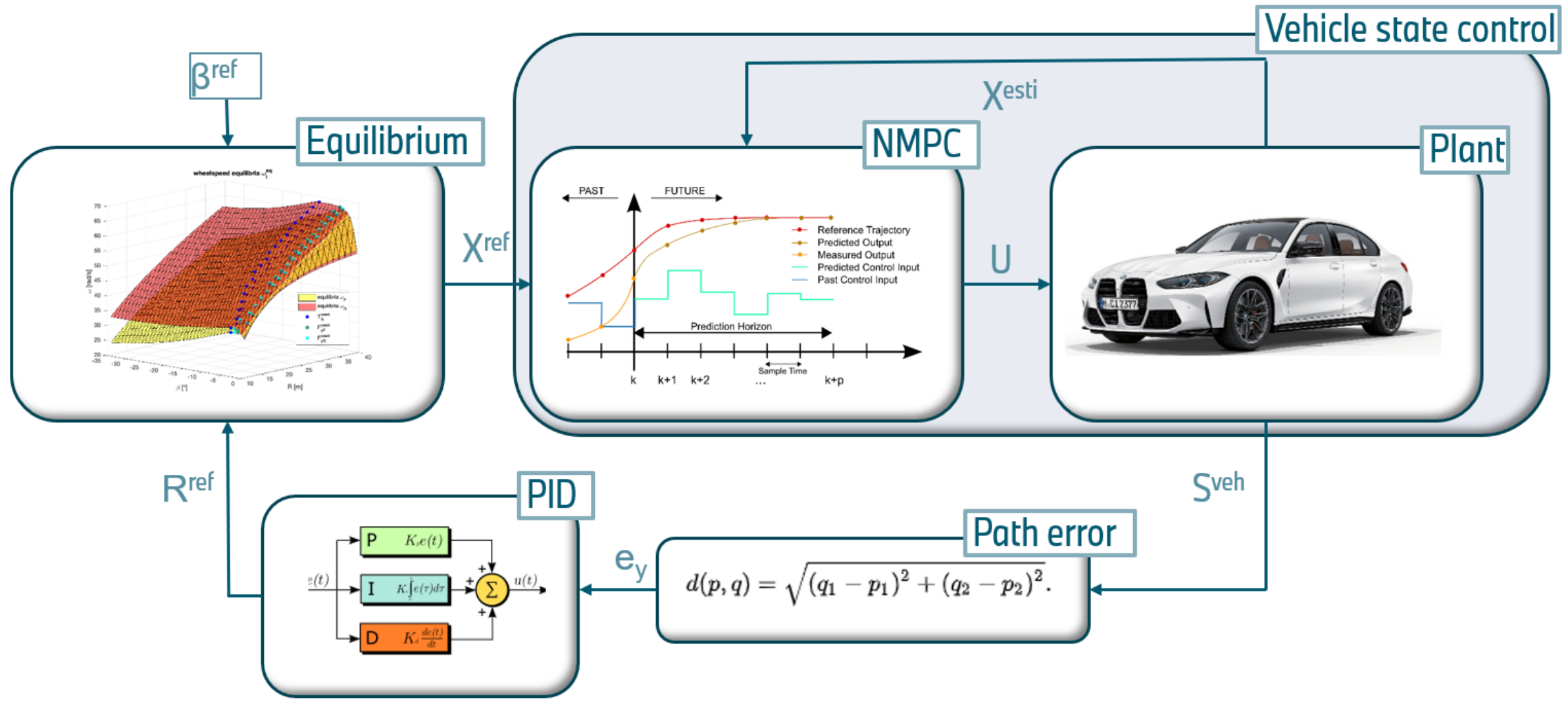}
    \caption{Architecture of the proposed controller.}
    \label{fig:MPC_total}
\end{figure}
The proposed controller architecture is shown in Fig. \ref{fig:MPC_total}, designed to stabilise the vehicle into a high sideslip angle and follow a desired path. The NMPC is used to compute the optimal control inputs to reach any feasible vehicle reference state. Thus, the NMPC is deployed to stabilise the vehicle in a high sideslip state and for conventional driving. Thus, it can bring the vehicle from conventional driving towards a drifting motion. The NMPC is based on the following online optimisation:
\begin{subequations}\label{eq:mpc}
\begin{align}
    \min_{\textbf{u} \in \mathbb{U}, \textbf{x}\in\mathbb{X}} \hspace{1em} & \sum_{k = 0}^N J(\textbf{x}_k, \textbf{u}_k)\\
        \textrm{s.t.} \hspace{2em} & \textbf{x}_0 = \textbf{x}_{\textrm{init}} \label{eq:init_condition} \\
& \textbf{x}_{k + 1} = f(\textbf{x}_k, \textbf{u}_k), \ k = 0, \hdots, N-1 \label{eq:dynamics}
\end{align}
\end{subequations}
where $J$ is the objective, which includes the quadratic errors between the defined equilibria and the vehicle sideslip angle and yaw rate states; furthermore, it includes a penalty on the steering rate $\dot{\delta}$. The latter is possible by introducing $\delta$ as an extra vehicle state correspondent to the integration of the $\dot{\delta}$. Thus, two benefits are counted: first, it is possible to add a constraint on $\dot{\delta}$, modelling the physical limitation of the steering actuator, and second, the control input $\delta$ becomes smoother. The $f(\textbf{x}_k, \textbf{u}_k)$ corresponds to the nonlinear single-track vehicle model. The optimisation problem in Eq.\ref{eq:mpc} is solved using the ACADO toolkit \cite{Houska2011a} with the Sequential Quadratic Programming solver at \SI{50}{Hz} with a prediction horizon $N = \SI{25}{}$.\compressParag

The desired equilibria are provided to the NMPC through dynamic referencing, which is implemented to influence the vehicle's direction. The offline computed equilibria are selected dynamically using $s_{x}$ to alter the path curvature such that the lateral error $e_y$ of a desired path is decreased. This is established by altering the desired path curvature in relation to the lateral error through a path-following controller that determines a compensating factor on the path curvature. The altered path curvature functions as an input for a lookup table that contains a complete set of equilibria, which, therefore, picks the optimal equilibrium that functions as a reference for the controller.\compressParag
\section{Simulation Results}
\label{sec:Results}
Subsection \ref{sub:SimulationSetting} describes how the proposed controller is evaluated in a high-fidelity environment, and subsection \ref{sub:SimulationResults} shows the proposed controller validation.

\subsection{Simulation Setup}
\label{sub:SimulationSetting}
The proposed controller is first evaluated in a simulation environment. The vehicle plant is a high-fidelity model of a BMW M3 Competition. It is a rear-wheel drive vehicle with a limited-slip differential (LSD), allowing for the locking rate modification.  The vehicle plant is based on a 17 degrees of freedom model, experimentally parameterised and validated by BMW. Magic Formula 5.2 is used to model the tyre dynamics. An alternating drifting manoeuvre on a high-friction surface is defined as a testing scenario. The manoeuvre is initialised with a high yaw rate state within friction limits, i.e. conventionally turning.
Furthermore, measurement noises and friction coefficient variations are included in the simulation. Friction coefficient variation is based on surface measurements at the BMW proving grounds of Aschheim, showing that seemingly homogeneous surfaces tend to have friction coefficient fluctuation of approximately 0.05. The amplitude and frequency of simulated measurement noises are based on experimental vehicle measurements.\compressParag

\subsection{Alternating Drifting Manoeuvre}
\label{sub:SimulationResults}
\begin{figure}[!t]
    \centering
    \includegraphics[width=0.85\linewidth]{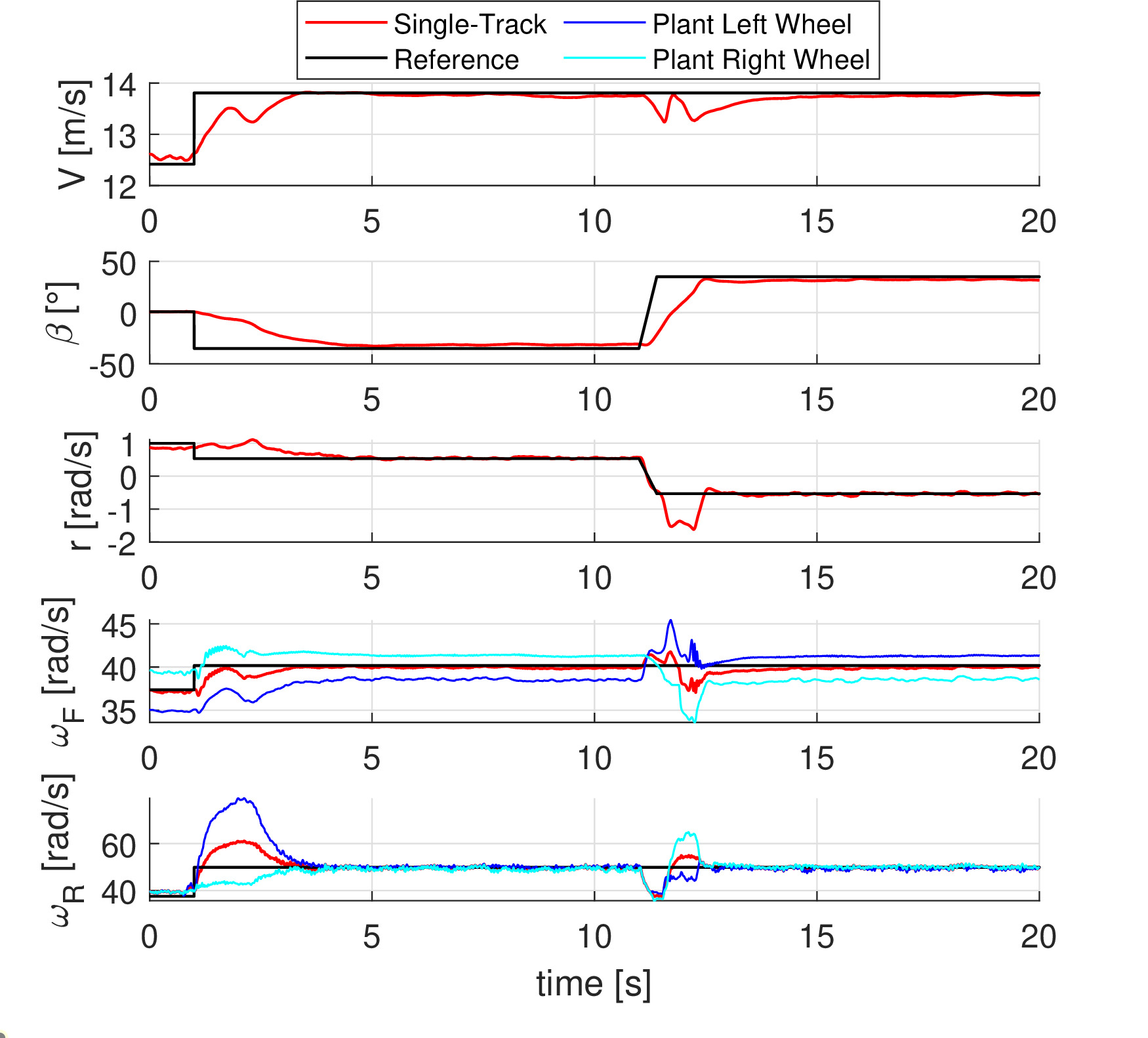}
    \caption{Vehicle states in the alternating drifting manoeuvre on a high friction surface.}
    \label{fig:ISARstates}
\end{figure}

\begin{figure}[!t]
    \centering
    \includegraphics[width=0.85\linewidth]{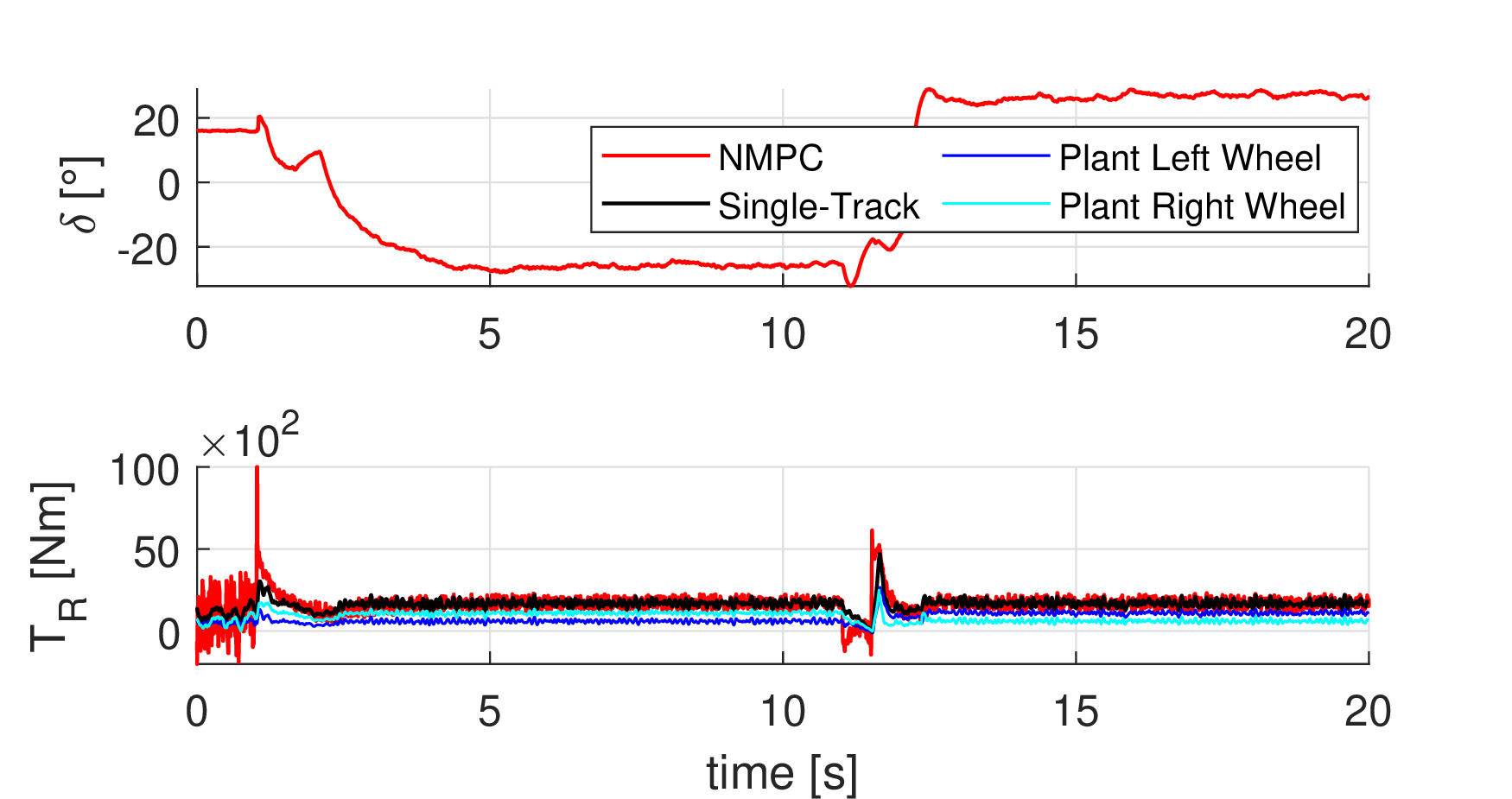}
    \caption{Control inputs in the alternating drifting manoeuvre on a high friction surface.}
    \label{fig:ISARinputs}
\end{figure}

The simulated vehicle states in the alternating drifting manoeuvre are shown in Fig. \ref{fig:ISARstates}. Once the drifting reference is initialised, the proposed controller can successfully track the initial left-hand drifting manoeuvre and perform a nearly instant transition into a right-hand drift with optimal tracking performance. In this case, the response of a \SI{30}{\%} LSD is reached as the left and right wheel speeds become similar, implying a locked differential state. Furthermore, the results highlight that the nonlinear single-track model captures the drifting vehicle dynamics, as the desired equilibrium is based upon that description.
The control inputs are shown in Fig. \ref{fig:ISARinputs}. Despite the same NMPC tuning for the complete simulation showing some erratic behaviour in the linear driving range,  the solver remained stable. The results highlight that the steering angle remains smooth, as the $\dot{\delta}$ is constrained in the optimisation. 
The simulator's response shows powertrain dynamics and systematic delays, as the single-track model calculation of the left and right wheel drive torque is not of equal peak magnitude. This implies that the rapid increase of desired drive torque is tracked slowly. However, the drifting states are reached as the solver computes the equilibrium drive torque.\compressParag
\section{Experimental Results}
Subsection \ref{sub:ExperimentalSetting} describes how the proposed controller is evaluated in the proving ground, and subsection \ref{sub:SemiResults} shows the experimental validation of the proposed NMPC for automated drifting.

\subsection{Experimental Setup}
\label{sub:ExperimentalSetting}
\begin{figure}[!t]
    \centering
    \includegraphics[width=1\linewidth,keepaspectratio]{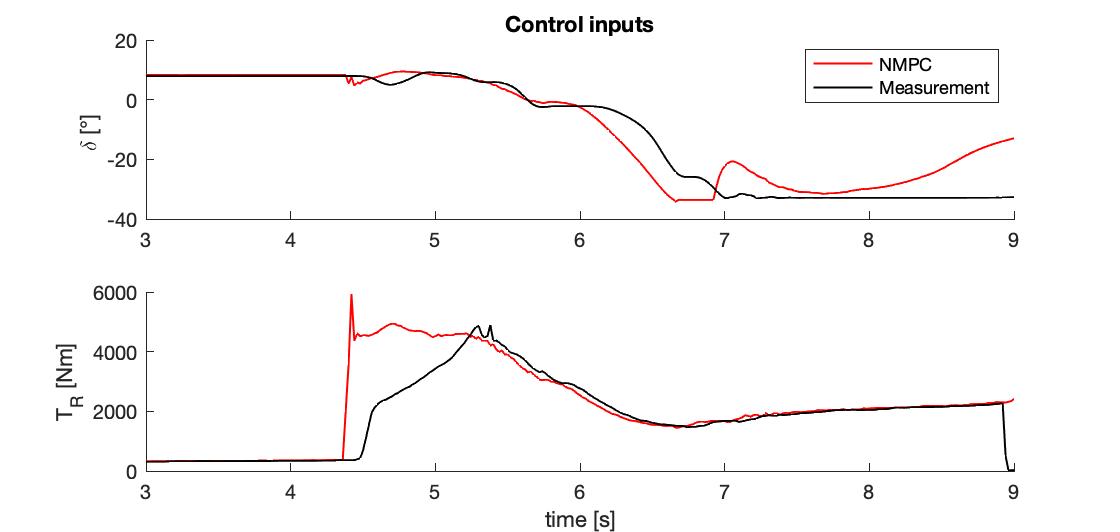}
        \caption{Desired and applied control inputs during automated drifting experiment.}
        \label{fig:MPC_delta_AS}
\end{figure}
The experimental validation is conducted with a BMW M3 Competition vehicle. The auto-generated real-time ACADO c-code is implemented on a dSPACE Autobox DS1007 platform, which also acts as an interface for desired Flexray and CAN bus communication protocols to send and receive information from the vehicle and desired external devices, e.g. GNNS or IMU. Measured and estimated state signals at \SI{10}{ms} are found on the vehicle Flexray. Regarding the control actuators, the human driver inputs provided to the driving assistance systems interface are replaced by the NMPC-optimised control inputs. However, the automated lane keeping, commonly available in a passenger vehicle, is not designed to be active in fully automated mode, so applied steering torque and angle through the driving assistance systems interface are greatly limited by safety features.\compressParag

When the proposed controller is validated, ignoring these limitations, it can only successfully bring the vehicle to a high sideslip angle condition but not stabilise it. Fig. \ref{fig:MPC_delta_AS} shows a deviation between the desired and measured steering angle for all performed tests reaching a high sideslip angle, where the front axle aligns with the direction of travel rather than compensates for the vehicle state. A possible explanation is the lack of applicable torque from the standard electric power steering motor that should be able to apply sufficient torque to overcome the self-aligning moment occurring at the front tyres. 
The e-motor can provide a maximum of \SI{\sim30}{Nm} of torque and is designed to function for several low-torque ADAS systems, such as automatic lane change. Thus,  the steering angle cannot be decreased as the NMPC desires and, as a result, the actual steering angle remains on its limit, which causes the vehicle to spin, as the front axle is not performing the desired compensation for the deviations in absolute velocity and vehicle sideslip angle.

Regarding these limitations, only software/hardware modifications or a steering robot can solve the steering angle limitation. Therefore, the proposed controller is validated with a semi-automated drifting manoeuvre. The proposed NMPC computes the optimal control inputs, but only the desired drive torque is provided to the vehicle while a human driver assists in the steering. In this experiment, we can validate the NMPC drive torque and compare the actual and desired steering input of the NMPC. The experiment is performed on a dry surface and a watered skid pad, respectively, with high and low friction conditions.

\subsection{Semi-Automated Drifting Manoeuvre}
\label{sub:SemiResults}
\begin{figure}[!t]
    \centering
    \includegraphics[width=0.9\linewidth,keepaspectratio]{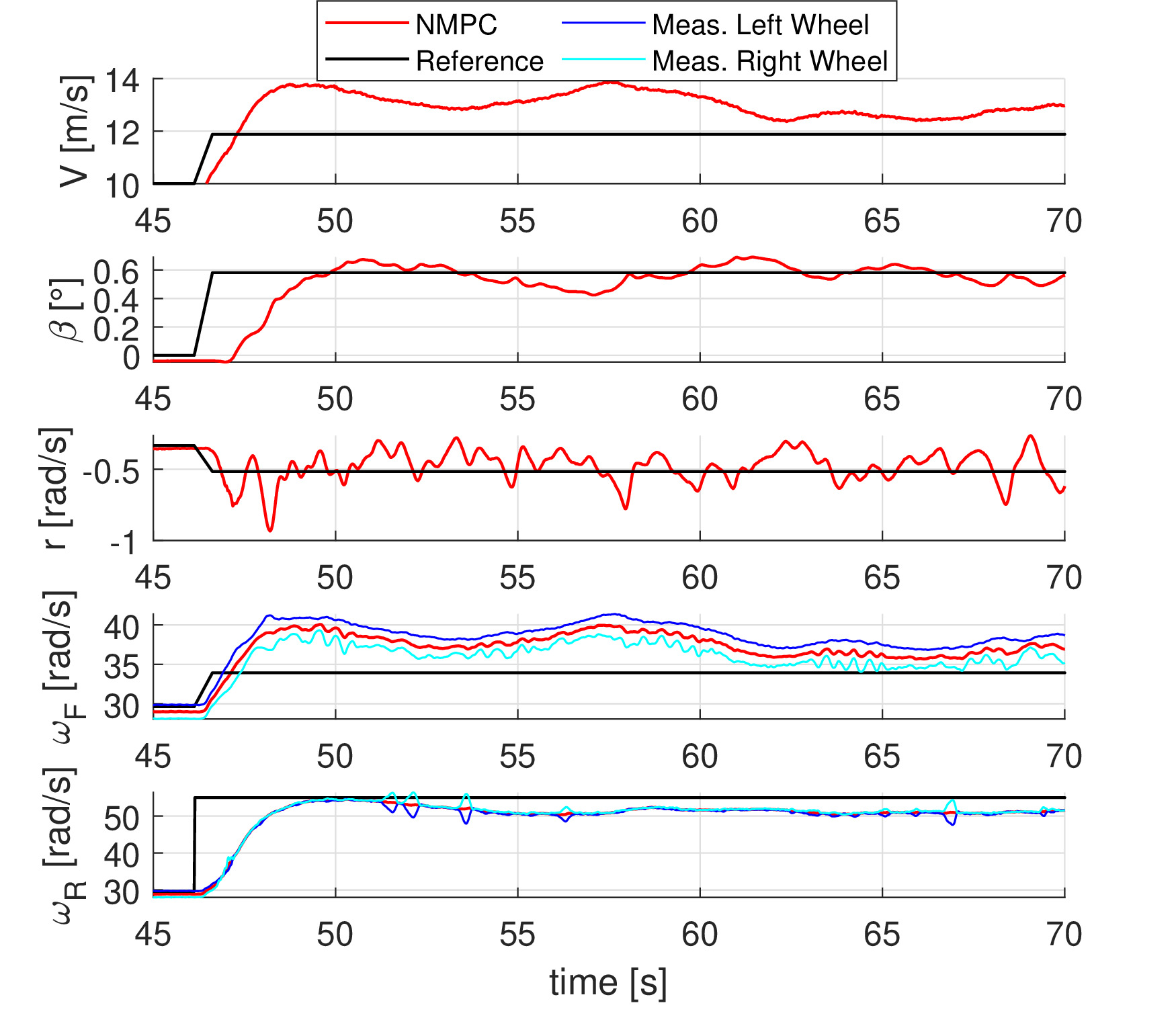}
    \caption{Vehicle states in semi-automated drifting manoeuvre.}
    \label{fig:MPC_Experiment_MS}
\end{figure}
\begin{figure}[!t] 
    \centering
    \includegraphics[width=0.9\linewidth, keepaspectratio]{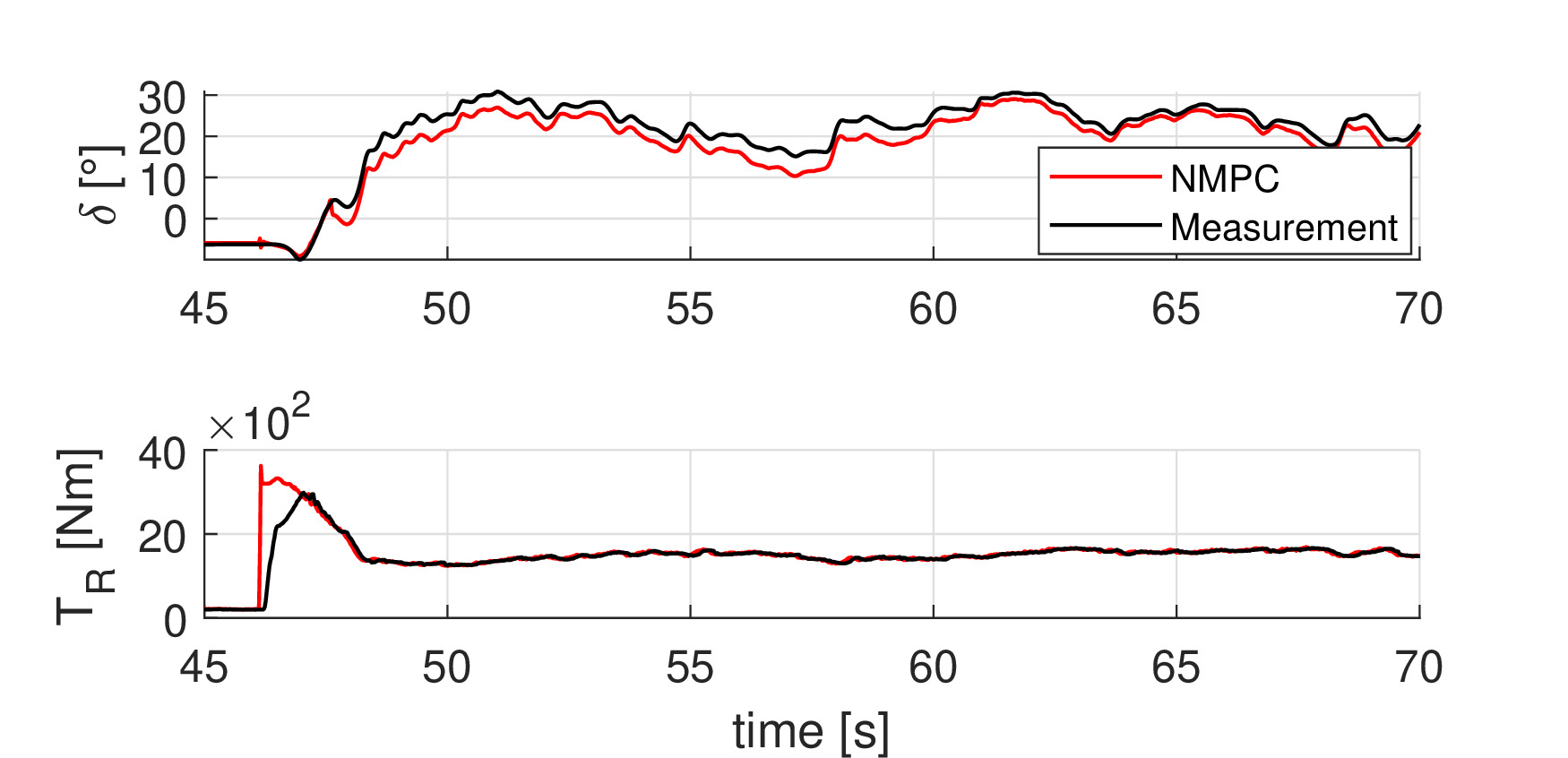}
    \caption{Actuated and optimised control inputs in a semi-automated drifting manoeuvre.}
    \label{fig:MPC_delta_MS}
\end{figure}
The vehicle states in a semi-automated drifting manoeuvre are shown in Fig. \ref{fig:MPC_Experiment_MS}. The vehicle successfully enters the desired drifting mode, and the high sideslip angle is stabilised for multiple rounds at \SI{30}{deg}. 
Thus, the NMPC driving torque is sufficient and solved optimally to bring the vehicle into drifting mode and maintain its steady-state drifting position. However, it is visible that the tracking performance of the velocity and front wheel speed is less accurate than the sideslip angle, yaw rate, and rear wheel speed. A possible explanation is that the controller tuning is optimised to bring the vehicle into a high sideslip angle, established with a high yaw rate and accurate rear wheel speed tracking. On the contrary, the front axle aligns with the direction of travel, and the front wheel speed and the absolute velocity are proportional under constant path curvature.

Fig. \ref{fig:MPC_delta_AS} shows the optimised and actuated control inputs. The comparison between steering angles demonstrates that the steering angle provided by the human driver is very similar to the desired steering angle optimised by the NMPC solver. The maximum variation in magnitude is \SI{\sim5}{deg}, demonstrating the NMPC capacity to stabilise the vehicle in drifting mode with appropriate actuators.
\section{Conclusions}
This paper presented a novel approach to automated drifting with a standard passenger vehicle, focused on a Nonlinear Model Predictive Control to stabilise and maintain the vehicle at high sideslip angle conditions. In this work, we experimentally verified the correspondence of the vehicle state equilibria computed with a nonlinear single-track model with the one measured in a proving ground. The experimental verification of the controller showed that using a standard production vehicle without significant hardware and software modifications is not possible due to the limited steering torque provided by the standard interface of electric power steering. However, a semi-automated drifting manoeuvre demonstrated the controller's capacity to bring a real vehicle into drifting mode and stabilise it at a high vehicle sideslip angle of \SI{30}{deg}. Future works involve performing the experimental validation of the proposed controller in a fully automated mode, using a steering robot as an actuator.
\addtolength{\textheight}{-5cm}   


\bibliographystyle{IEEEtran}
\bibliography{IEEEabrv,references}

\end{document}